\documentclass{article}

\usepackage{microtype}
\usepackage{graphicx}
\usepackage{booktabs} 
\usepackage{amsmath,amssymb}
\usepackage[dvipsnames]{xcolor}
\usepackage{graphicx}
\usepackage{bm}
\usepackage{subcaption}
\usepackage[export]{adjustbox}
\usepackage{tikz}
\usetikzlibrary{arrows, arrows.meta, shapes, calc, intersections, positioning, decorations.pathreplacing, fit, backgrounds, patterns}

\usepackage[hyphens]{url}

\usepackage{hyperref}
\usepackage[noabbrev,nameinlink]{cleveref}

\newcommand{\x}{\mathbf{x}}
\newcommand{\y}{\mathbf{y}}

\newcommand{\UU}{\bar{\mathbf{U}}}
\newcommand{\LL}{\mathbf{L}}
\newcommand{\bmu}{\boldsymbol\mu}
\newcommand{\bsigma}{\bar{\boldsymbol\sigma}}
\newcommand{\bSigma}{\boldsymbol\Sigma}

\DeclareMathSymbol{\sminus}{\mathbin}{AMSa}{"39}

\definecolor{vll-orange}{HTML}{E37238}
\definecolor{vll-green}{HTML}{96BF0D}
\definecolor{vll-dark}{HTML}{464646}
\definecolor{vll-light}{HTML}{757575}

\usepackage[accepted]{icml2020}

\icmltitlerunning{Technical report: Training Mixture Density Networks with full covariance matrices}

\begin{document}

\twocolumn[
\icmltitle{Training Mixture Density Networks with full covariance matrices}

\icmlsetsymbol{equal}{*}
\begin{icmlauthorlist}
    \icmlauthor{Jakob Kruse}{vll}
\end{icmlauthorlist}
\icmlaffiliation{vll}{Visual Learning Lab, Heidelberg University. March 2020}
\icmlcorrespondingauthor{jakob.kruse@iwr.uni-heidelberg.de}
\icmlkeywords{Machine Learning, Invertible Neural Networks, Normalizing Flow, Fourier Curve, ICML}
\vskip 0.3in
]
\printAffiliationsAndNotice{}  

\begin{abstract}

Mixture Density Networks are a tried and tested tool for modelling conditional probability distributions.
As such, they constitute a great baseline for novel approaches to this problem.
In the standard formulation, an MDN takes some input and outputs parameters for a Gaussian mixture model with restrictions on the mixture components' covariance.
Since covariance between random variables is a central issue in the conditional modeling problems we were investigating, I derived and implemented an MDN formulation with unrestricted covariances.
It is likely that this has been done before, but I could not find any resources online.
For this reason, I have documented my approach in the form of this technical report, in hopes that it may be useful to others facing a similar situation.

\end{abstract}

\section{Preliminaries}

The standard normal -- or Gaussian -- distribution acts as the foundation for countless modelling approaches in statistics and machine learning.
As opposed to a single Gaussian, which is a very limited model, mixtures of multivariate Gaussians can represent or approximate almost any density of interest while remaining intuitive and easy to handle.

\subsection{Gaussian mixture model}

The density at point $\x$ under an $N$-dimensional Gaussian mixture model with $K$ components is a weighted sum of the densities under each Gaussian component:

\begin{align}
    p(\x)
    &= \sum_{i=1}^K \omega_i \cdot p_i(\x) \\
    &= \sum_{i=1}^K \omega_i \cdot \mathcal{N}(\x \,|\, \bmu_i, \bSigma_i) \nonumber\\
    &= \sum_{i=1}^K \omega_i \cdot \frac{\exp{\left( \sminus\tfrac{1}{2} (\x - \bmu_i)^\top \cdot \bSigma_i^{\sminus 1} \cdot (\x - \bmu_i) \right)}}{\sqrt{\left( 2\pi \right)^K \cdot |\bSigma_i|}} \nonumber
\end{align}
We can leave out the constant factor and get
\begin{align}
    p(\x)
    &\propto \sum_{i=1}^K \omega_i \cdot \frac{\exp{\left( \sminus\tfrac{1}{2} (\x - \bmu_i)^\top \cdot \bSigma_i^{\sminus 1} \cdot (\x - \bmu_i) \right)}}{\sqrt{|\bSigma_i|}} , \nonumber\\
    &\text{all with component weights} \sum_{i=1}^K \omega_i = 1.
\end{align}
Each component has a mean $\bmu_i$ and a covariance matrix $\bSigma_i$.
A common simplification is to restrict the covariance matrices to be diagonal,
in which case the determinant required in the denominator reduces to the product of the diagonal entries.

Since these diagonal entries must be strictly positive, a convenient parameterization is $\bSigma_i^{\sminus1} = \bsigma_i^\top \cdot \mathbf{I}_N \cdot \bsigma_i$ with $(\bsigma_i)_j = \exp{(\boldsymbol\sigma_i)_j}$ for some unconstrained vector $\boldsymbol\sigma_i$.
With this the denominator just becomes the product
\begin{align}
    \frac{1}{\sqrt{|\bSigma_i|}}
    &= \frac{1}{\sqrt{\prod_{j=1}^N (\bsigma_i)^{-2}_j}}
    = \prod_{j=1}^N (\bsigma_i)_j \nonumber
\end{align}
and the above density can be expressed as
\begin{align}
    p(\x)
    &\propto \sum_{i=1}^K \omega_i \cdot \exp{\left( \sminus\tfrac{1}{2} \|(\x - \bmu_i) \odot \bsigma_i\|_2^2 \right)} \cdot \prod_{j=1}^N (\bsigma_i)_j
\end{align}

For numerical stability we can compute the contribution of each single component $p_i(\x)$ as a sum in log-space as
\begin{align}
    \log p_i(\x)
    &\propto \sminus\tfrac{1}{2} \|(\x - \bmu_i) \odot \bsigma_i\|_2^2 + \sum_{j=1}^N \log \ (\bsigma_i)_j \nonumber\\
    &\propto \sminus\tfrac{1}{2} \|(\x - \bmu_i) \odot \bsigma_i\|_2^2 + \sum_{j=1}^N (\boldsymbol\sigma_i)_j
\end{align}

\subsection{Mixture Density Networks}

A Mixture Density Network \cite{bishop1994mixture}, or MDN for short, is a neural network that outputs parameters $\omega_{i|\theta}, \bmu_{i|\theta}, \boldsymbol\sigma_{i|\theta}$ for all components $i$ of a Gaussian mixture model $p_\theta$ based on some input $\y$, where $\theta$ are the trainable network weights.
As such, an MDN parameterizes the conditional density $p_\theta(\x \,|\, \y)$.

The standard objective for training an MDN is maximum likelihood of a training set $X,Y$ under $p_\theta(\x \,|\, \y)$.
To this end, we minimize the \emph{negative log-likelihood} with respect to the network parameters $\theta$:
\begin{align}
    \mathcal{L}(\theta)
    &= \mathbb{E}_{\x \in X, \y \in Y} \big[ \sminus \log p_\theta(\x \,|\, \y) \big] \label{eq:nll-loss}\\
    &\propto \mathbb{E}_{\x \in X, \y \in Y} \Biggl[ \sminus \log
    \sum_{i=1}^K \omega_{i|\theta} \cdot \exp \Big( \!\sminus\! \tfrac{1}{2} \big\|(\x \sminus \bmu_{i|\theta}) \nonumber\\[-4pt]
    &\hspace{7em} \odot \bsigma_{i|\theta}\big\|_2^2
    + \sum_{j=1}^N (\boldsymbol\sigma_{i|\theta})_j \Big) \Biggr]
    \intertext{Using Jensen's inequality, we can pull the logarithm into the sum.
    This yields an upper bound on the negative log-likelihood with more robust behavior especially during early training, when the predicted parameters are still somewhat volatile:}
    \mathcal{L}(\theta)
    &\leq \mathbb{E}_{\x \in X, \y \in Y} \Biggl[ \sminus
    \sum_{i=1}^K \Big( \log \omega_{i|\theta} - \tfrac{1}{2} \big\|(\x \sminus \bmu_{i|\theta}) \nonumber\\[-4pt]
    &\hspace{7em} \odot \bsigma_{i|\theta}\big\|_2^2 + \sum_{j=1}^N (\boldsymbol\sigma_{i|\theta})_j \Big) \Biggr]
\end{align}

We can minimize this loss function via back-propagation and standard gradient descent methods for deep learning.

\subsection{Sampling}

To sample one value $\x$ from a Gaussian mixture, we first pick a mixture component $i$ with probability proportional to its weight $\omega_i$ and then draw $\x$ from $p_i(\x)$.
In practice and in the case of diagonal covariance matrices, this means
\begin{align}
    \x
    &\sim \mathcal{N}(\x \,|\, \bmu_i, \bSigma_i) \nonumber\\
    &= \bmu_i + \boldsymbol\eta \oslash \bsigma_i \text{ \ with } \boldsymbol\eta \sim \mathcal{N}\left( \boldsymbol\eta \,|\, \mathbf{0}, \mathbf{I}_N \right), \nonumber
\end{align}
where $\bmu_i$ and $\bsigma_i$ can be direct outputs of an MDN for all $i$.

\section{Full covariance matrices}

\begin{figure}[t]
\centering
\includegraphics[width=\columnwidth]{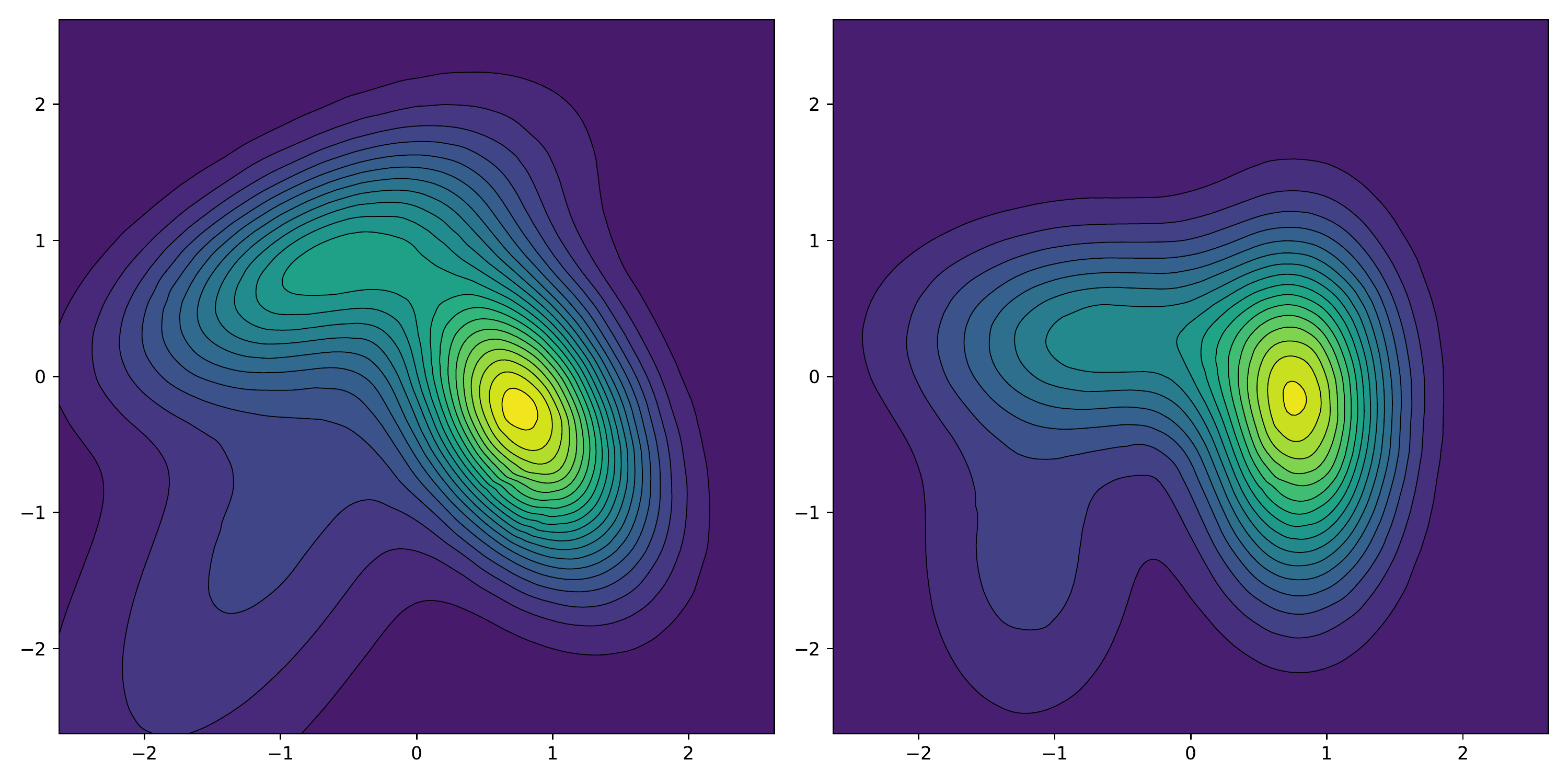}%
\vspace{-1em}%
\caption{Difference between a Gaussian mixture with full covariances \emph{(left)} and with diagonal covariances, i.e.~axis-parallel \emph{(right)}}%
\label{fig:gmm-full-vs-diagonal}
\end{figure}

A multivariate Gaussian with strictly diagonal covariance matrix, as we have used above for the mixture components, has strong limitations as to what densities can be modelled.
\Cref{fig:gmm-full-vs-diagonal} shows this for three components in 2d space:
while components of the unconstrained mixture on the left can be oriented arbitrarily in the plane, components of the diagonal mixture on the right can only be scaled along the axes of the coordinate system.

We want to make use of the much greater flexibility of the unconstrained model, but avoid the costly computation of full matrix inverses and determinants, as well as covariance matrices that do not describe valid Gaussian densities.
In the following, we will therefore show a parameterization that guarantees a valid density while offering (relatively) efficient training and sampling.

\subsection{Parameterization}

If we look at the general density formula for a single mixture component $p_i(\x)$, we can pull the covariance matrix $\bSigma_i$ from the denominator and rewrite as:
\begin{align}
    p_i(\x)
    &\propto \frac{\exp{\left( \sminus\tfrac{1}{2} (\x \sminus \bmu_i)^\top \bSigma_i^{\sminus 1} (\x \sminus \bmu_i) \right)}}{\sqrt{|\bSigma_i|}} \nonumber\\
    &\propto \exp{\left( \sminus\tfrac{1}{2} (\x \sminus \bmu_i)^\top \bSigma_i^{\sminus 1} (\x \sminus \bmu_i) \right)} \cdot |\bSigma_i|^{\sminus\frac{1}{2}} \nonumber\\
    &\propto \exp{\left( \sminus\tfrac{1}{2} (\x \sminus \bmu_i)^\top \bSigma_i^{\sminus 1} (\x \sminus \bmu_i) \right)} \cdot |\bSigma_i^{\sminus 1}|^\frac{1}{2}
    \label{eq:pi-only-inverse}
\end{align}
Now $\bSigma_i$ only occurs in the equation as its inverse, the \emph{precision matrix} $\bSigma_i^{\sminus 1}$.

For a valid multivariate Gaussian, $\bSigma_i$ and $\bSigma_i^{\sminus 1}$ must be positive-definite matrices.
We can thus characterize the precision matrix $\bSigma_i^{\sminus 1}$ by its Cholesky decomposition using an upper triangular matrix $\UU_i$ with strictly positive diagonal entries:
\begin{align}
    \bSigma_i^{\sminus 1}
    &= \UU_i^\top \UU_i
\end{align}
Once again we can use the exponential function to enforce positivity of the diagonal, by taking
\begin{align}
    (\UU_i)_{jk} &=
    \begin{cases}
        (\mathbf{U}_i)_{jk},& \text{if } j \neq k \\
        \exp \, (\mathbf{U}_i)_{jk},& \text{otherwise}
    \end{cases}
\end{align}
for an unconstrained upper triangular matrix $\mathbf{U}_i$ predicted by the MDN.

This offers an efficient way to compute the matrix determinant in \cref{eq:pi-only-inverse} as the product of the diagonal entries of the Cholesky factor $\UU_i$:

\begin{align}
    \left| \bSigma_i^{\sminus 1} \right|
    &= \left| \UU_i^\top \UU_i \right| \nonumber\\
    &= \left| \UU_i^\top \right| \cdot \left| \UU_i \right| \nonumber\\
    &= \prod_{j=1}^N \text{diag}\big( \UU_i^\top \big)_j \cdot \prod_{j=1}^N \text{diag}\big( \UU_i \big)_j \nonumber\\
    &= \left( \prod_{j=1}^N \text{diag}\big( \UU_i \big)_j \right)^{\!2} \nonumber\\
    \left| \bSigma_i^{\sminus 1} \right|^\frac{1}{2}
    &= \prod_{j=1}^N \text{diag}\big( \UU_i \big)_j \\
    \intertext{Its numerically more stable logarithm has the form}
    \log \left| \bSigma_i^{\sminus 1} \right|^\frac{1}{2}
    &= \sum_{j=1}^N \log \text{diag}\big( \UU_i \big)_j \nonumber\\
    &= \sum_{j=1}^N \text{diag}\big( \mathbf{U}_i \big)_j \,.
\end{align}

With this, we can express the log-density under one mixture component $p_i(\x)$ as
\begin{align}
    \log p_i(\x)
    &\propto \sminus\tfrac{1}{2} (\x \!\sminus\! \bmu_i)^\top \bSigma_i^{\sminus 1} (\x \!\sminus\! \bmu_i) + \log |\bSigma_i^{\sminus 1}|^\frac{1}{2} \nonumber\\
    &\propto \sminus\tfrac{1}{2} (\x \!\sminus\! \bmu_i)^\top \UU_i^\top \cdot \UU_i (\x \!\sminus\! \bmu_i) + \sum_{j=1}^N \text{diag}\big( \mathbf{U}_i \big)_j \nonumber\\
    &\propto \sminus\tfrac{1}{2} \left\| \UU_i (\x \!\sminus\! \bmu_i) \right\|_2^2 + \sum_{j=1}^N \text{diag}\big( \mathbf{U}_i \big)_j \ ,
\end{align}
which is remarkably similar to the max-likelihood loss used to train normalizing flow networks.

Note that the number of parameters the MDN has to predict grows linearly with the number $K$ of mixture components, but quadratically with the number of dimensions $N$
since the triangular matrix $\mathbf{U}_i$ must be populated for each component.

\subsection{Training}

With the above parameterization, we need the MDN to output parameters $\omega_{i|\theta}, \bmu_{i|\theta}$ and $\mathbf{U}_{i|\theta}$ for each mixture component $p_{i|\theta}(\x)$.
We optimize the network weights $\theta$ according to the same maximum likelihood criterion used in \cref{eq:nll-loss}, which now takes the form
\begin{align}
    \mathcal{L}(\theta)
    &= \mathbb{E}_{\x \in X, \y \in Y} \big[ \sminus \log p_\theta(\x \,|\, \y) \big] \nonumber\\
    &\propto \mathbb{E}_{\x \in X, \y \in Y} \Biggl[ \sminus \log
    \sum_{i=1}^K \omega_{i|\theta} \cdot \exp \Big( \sum_{j=1}^N \text{diag}\big( \mathbf{U}_{i|\theta} \big)_j
    \nonumber\\[-4pt]
    &\hspace{7em} - \tfrac{1}{2} \left\| \UU_{i|\theta} (\x \!\sminus\! \bmu_{i|\theta}) \right\|_2^2 \Big) \Biggr] . \\
    \intertext{Jensen's inequality again gives us an upper bound to be used in the early training regime:}
    \mathcal{L}(\theta)
    &\leq \mathbb{E}_{\x \in X, \y \in Y} \Biggl[ \sminus
    \sum_{i=1}^K \Big( \log \omega_{i|\theta} + \sum_{j=1}^N \text{diag}\big( \mathbf{U}_{i|\theta} \big)_j
    \nonumber\\[-4pt]
    &\hspace{7em} - \tfrac{1}{2} \left\| \UU_{i|\theta} (\x \!\sminus\! \bmu_{i|\theta}) \right\|_2^2 \Big) \Biggr]
\end{align}

\subsection{Sampling}

After running the network to obtain parameters $\omega_i, \bmu_i, \mathbf{U}_i$ for a Gaussian mixture model, we start by choosing one component $i$ as described earlier.
Samples from a multivariate Gaussian with full covariance are drawn as
\begin{align}
    \x
    &\sim \mathcal{N}(\x \,|\, \bmu_i, \bSigma_i) \nonumber\\
    &= \bmu_i + \LL_i \cdot \boldsymbol\eta \nonumber
\end{align}
with $\boldsymbol\eta \sim \mathcal{N}\left( \boldsymbol\eta \,|\, \mathbf{0}, \mathbf{I}_N \right)$ and $\LL_i$ chosen to factorize the covariance matrix such that $\bSigma_i = \LL_i \LL_i^\top$.

Noting that the parameterization we have introduced for our network already gives us the Cholesky root $\UU_i$ of the precision matrix, i.e.~$\bSigma_i^{\sminus1} = \UU_i^\top \UU_i$, we can use inverse of its transpose $\LL_i = \UU_i^{\sminus \top}$ to factorize the covariance matrix\footnote{\url{https://math.stackexchange.com/a/2489142}}.
Since $\UU_i^\top$ is triangular, this inversion can be performed efficiently via back substitution.

\section{Implementation}

An implementation of the model described above using \texttt{pytorch} can be found in our Framework for Easily Invertible Architectures,
\texttt{FrEIA}\footnote{\url{https://github.com/VLL-HD/FrEIA}}, under the module name \texttt{GaussianMixtureModel}.

In the context of this framework, the GMM acts as an invertible building block which maps between a data point $\x$ and its latent code $\boldsymbol\eta$.
Invertibility is given either by specifying (or reproducibly sampling) a fixed component index $i$, or by extending the map to consider all mixture components simultaneously, which is also needed for training.

The GMM block has no parameters of its own, but instead takes $\boldsymbol\omega \in \mathbb{R}^{b \times K}$, $\bmu \in \mathbb{R}^{b \times K \times N}$, $\mathbf{U} \in \mathbb{R}^{b \times K \times N(N-1)/2}$ and the optional index $i$ as conditional inputs.
The values $\boldsymbol\omega, \bmu$ and $\mathbf{U}$ should come from a feed-forward network taking $\y$ as input.
Then the latter network can be trained via back-propagation through the GMM block, using the negative log-likelihood loss function which is also supplied.

\bibliography{references}
\bibliographystyle{icml2020}

\end{document}